\pdfoutput=1
\documentclass[11pt]{article}
\usepackage[]{naacl2021}
\usepackage{times}
\usepackage{latexsym}
\usepackage{booktabs}
\usepackage{graphicx}
\usepackage{siunitx}
\usepackage{graphics}
\usepackage{multirow}
\usepackage{url}
\usepackage{amsmath,amssymb,amsopn,amsthm}
\usepackage{algorithm,algpseudocode,algorithmicx}

\usepackage[T1]{fontenc}
\usepackage{subcaption}
\usepackage[utf8]{inputenc}

\usepackage{microtype}

\usepackage[most]{tcolorbox}
\usepackage{array}

\title{Revisiting the Weaknesses of Reinforcement Learning \\for Neural Machine Translation}

\author{Samuel Kiegeland \\
  Heidelberg University \\
  \texttt{kiegeland@cl.uni-heidelberg.de} \\\And
  Julia Kreutzer \\
  Google Research \\
  \texttt{jkreutzer@google.com} \\}

\begin{document}
\maketitle
\begin{abstract}
Policy gradient algorithms have found wide adoption in NLP, but have recently become subject to criticism, doubting their suitability for NMT. \citet{choshen} identify multiple weaknesses and suspect that their success is determined by the shape of output distributions rather than the reward. In this paper, we revisit these claims and study them under a wider range of configurations. Our experiments on in-domain and cross-domain adaptation reveal the importance of exploration and reward scaling, and provide empirical counter-evidence to these claims.

\end{abstract}

\section{Introduction}
In neural sequence-to-sequence learning, in  particular Neural Machine Translation (NMT), Reinforcement Learning (RL) has gained attraction due to the suitability of Policy Gradient (PG) methods for the end-to-end training paradigm
\citep{RanzatoETAL:16, li-etal-2016-deep, 
yu-etal-2017-learning,
li-etal-2018-paraphrase,
flachs-etal-2019-historical, sankar-ravi-2019-deep}.
The idea is to let the model explore the output space beyond the reference output that is used for standard cross-entropy minimization, by reinforcing model outputs according to their quality, effectively increasing the likelihood of higher-quality samples. The classic exploration-exploitation dilemma from RL is addressed by sampling from a pretrained model's softmax distribution over output tokens, such that the model entropy steers exploration.

For the application of NMT, it was firstly utilized to bridge the mismatch between the optimization for token-level likelihoods during training and the corpus-level held-out set evaluations with non-differentiable/decomposable metrics like BLEU 
\citep{RanzatoETAL:16,edunov-etal-2018-classical}, and secondly to reduce \emph{exposure bias} in autoregressive sequence generators \citep{RanzatoETAL:16,Wang2020OnEB}. It has furthermore been identified as a promising tool to adapt pretrained models to new domains or user preferences by replacing reward functions with human feedback in human-in-the-loop learning \citep{SokolovETALnips:16, nguyen-etal-2017-reinforcement}.

Recently, the effectiveness of these methods has been questioned: \citet{choshen} identify multiple theoretical and empirical weaknesses, leading to the suspicion that performance gains with RL in NMT are not due to the reward signal. The most surprising result is that the replacement of a meaningful reward function (giving higher rewards to higher-quality translations) by a constant reward (reinforcing all model samples equally) yields similar improvements in BLEU.  
To explain this counter-intuitive result, \citet{choshen} conclude that 
a phenomenon called the \emph{peakiness effect} must be responsible for performance gains instead of the reward. This means that the most likely tokens in the beginning gain probability mass regardless of the rewards they receive during RL training.
If this hypothesis was true, then the perspectives for using methods of RL for encoding real-world preferences into the model would be quite dire, as models would essentially be stuck with whatever they learned during supervised pretraining and not reflect the feedback they obtain later on. 

However, the analysis by \citet{choshen} missed a few crucial aspects of RL that have led to empirical success in previous works: First, \emph{variance reduction} techniques such as the average reward baseline were already proposed with the original Policy Gradient by \citet{williams}, and proved effective for NMT \citep{kreutzer-etal-2017-bandit, nguyen-etal-2017-reinforcement}. Second, the \emph{exploration-exploitation trade-off} can be controlled by modifying the sampling function \citep{sharaf-daume-iii-2017-structured}, which in turn influences the peakiness. 

We therefore revisit the previous findings with NMT experiments differentiating model behavior between in-domain and out-of-domain adaptation, controlling exploration, reducing variance, and isolating the effect of reward scaling. This allows us to establish a more holistic view of the previously identified weaknesses of RL. In fact, our experiments reveal that \textbf{improvements in BLEU can not solely be explained by increased peakiness}, and that simple methods encouraging \textbf{stronger exploration can successfully move previously lower-ranked token into higher ranks}. We observe generally low empirical gains in in-domain adaptation, which might explain the surprising success of constant rewards in \citet{choshen}. However, we find \textbf{that rewards and their scaling do matter for domain adaptation}. Furthermore, our results corroborate the auspicious findings of ~\citet{Wang2020OnEB} that RL mitigates exposure bias. Our paper thus reinstates the potential of RL for  model adaptation in NMT, and puts previous pessimistic findings into perspective. The code for our experiments is publicly available.\footnote{\url{https://github.com/samuki/reinforce-joey}}


\section{RL for NMT}
The objective of RL in NMT is to maximize the expected reward for the model's outputs 
with respect to the parameters $\theta$: $\arg\max_{\theta}\mathbb{E}_{p_{\theta}(y\mid x)}[\Delta(y, y^{\prime})]$.
where $y'$ denotes a reference translation, $y$ is the generated translation and $\Delta$ is a metric (e.g. BLEU \citep{papineni2002bleu}), rewarding similarities to the reference. Applying the log derivative trick, the following gradient 
can be derived: 
\begin{equation}
    \nabla_{\theta}  = \mathbb{E}_{p_{\theta}(y\mid x)}[\Delta(y, y^{\prime})\nabla_{\theta}\log p_{\theta}(y\mid x)].
    \label{eqn:gradient}
\end{equation}
The benefit of Eq.~\ref{eqn:gradient} is that it does not require differentiation of $\Delta$ which allows for direct optimization of the BLEU score or human feedback.\footnote{Rewards may be obtained without reference translations $y^{\prime}$, hence $\Delta(y)$ can replace $\Delta(y, y^{\prime})$ in the following equations.}

\subsection{Policy Gradient}
However, computing the gradient requires the summation over all $y \in \mathcal{V}_{trg}^{m}$, which is computationally infeasible for large sequence lengths $m$ and vocabulary sizes $\mathcal{V}_{trg}$ as they are common in NMT.
 Therefore, Eq.~\ref{eqn:gradient} is usually approximated through Monte Carlo sampling~\citep{williams} resulting in unbiased estimators of the full gradient. 

 We draw one sample from the multinomial distribution defined by the model's softmax to approximate Eq.~\ref{eqn:gradient} \citep{RanzatoETAL:16, kreutzer-etal-2017-bandit, choshen}, which results in the following update rule with learning rate $\alpha$: 
\begin{align}\label{eq:pg-update}
    & u_{k} = \nabla_{\theta}\log p_{\theta}(y\mid x)\Delta(y, y^{\prime})\\ 
    & \theta_{t+1} = \theta_{t} + \alpha u_{k} 
\end{align}

\subsection{Softmax Temperature}
The temperature $\tau$ of the softmax distribution $\exp(y_i/\tau)/\sum_j \exp(y_j/\tau)$ can be used to control the amount of exploration during learning. Setting $0<\tau<1$ results in less diverse samples while setting $\tau>1$ increases the diversity and also the entropy of the distribution. Lowering the temperature (i.e. making the distribution peakier) may be used to make policies more deterministic towards the end of training \citep{sutton1998rli, Rose:98, SokolovETAL:17}, while we aim to reduce peakiness by increasing the temperature.

\begin{table*}[t]
    \centering
    \begin{tabular}{l|cc|ccc}
        \toprule
 \textbf{Model}   & $\Delta p_{top10}$ & $\Delta p_{mode}$ & BLEU ($k=1$) & BLEU ($k=5$) & BLEU ($k=50$)\\
 \midrule
Pretraining (IWSLT14) & 0 & 0 & 33.49  & 34.12 & 33.88\\
\midrule
PG ($n=1$) & 11.46 & 24.82 & $33.84 \pm 0.05$ & $34.24 \pm 0.04$ & $34.16 \pm 0.13$ \\ 
PG + scaled  & 11.36 & 24.42 & $33.91 \pm  0.14$ & $34.30 \pm 0.10$  & $34.19 \pm 0.11$ \\
PG + average bl  & 12.54 & 27.84 & $34.20 \pm 0.04$ & $34.40 \pm 0.03$ & $34.30 \pm 0.04$  \\ 
PG + $\tau = 1.2$  & 6.08  & 16.91 & $33.88 \pm 0.04$  & $34.15 \pm 0.02$ &  $34.11 \pm 0.01$ \\
PG + $\tau = 0.8$  & 14.29  & 29.74 & $33.80 \pm 0.03$ & $34.26 \pm 0.11$ & $34.14 \pm 0.10$  \\
PG + constant  & 1.42  & 1.02 & $33.53 \pm 0.04$ & $34.13 \pm 0.01$ & $33.91 \pm 0.03$ \\
 \midrule
 PG + average bl + $\tau = 1.05$  & 12.09  & 27.51  & $34.37 \pm 0.06$ & $34.51 \pm 0.11$ & $34.46 \pm 0.09$ \\ 
 PG + average bl + $\tau = 0.95$  &  13.36 & 29.83 &  $34.28 \pm  0.10$ & $34.51 \pm  0.06$ & $34.41 \pm 0.09$   \\
\midrule 
MRT ($n=5$) & 12.93 & 32.85 & $34.52 \pm 0.06$ & $34.68 \pm 0.05$ & $34.63 \pm 0.05$ \\
\bottomrule
    \end{tabular}
    \caption{In-domain adaptation: Peakiness indicators (\%), and IWSLT14 test set results for beam size $k$.}
    \label{tab:in_domain}
\end{table*}

\subsection{Modified Rewards}
\label{sec:modidied rewards}
Variance reduction techniques were already suggested by \citet{williams} and found to improve generalization for NMT \citep{kreutzer-etal-2017-bandit}. The simplest option is the \emph{ baseline reward}, which in practice is realized by subtracting a running average of historic rewards from the current reward $\Delta$ in Eq.~\ref{eq:pg-update}. It represents an expected reward, so that model outputs get more strongly reinforced or penalized if they diverge from it.

In addition to variance reduction, subtracting baseline rewards also change the scale of rewards (e.g. $\Delta \in [0,1]$ for BLEU becomes $\Delta \in [-0.5, 0.5]$), allowing updates towards or away from samples by switching the sign of $u_k$ (Eq.~\ref{eq:pg-update}). The same range of rewards can be obtained by \emph{re-scaling} them, e.g., to
$\frac{\Delta(y, y')-\min}{\max-\min} -0.5$ with the minimum ($\min$) and maximum ($\max$) $\Delta$ within each batch.
\subsection{Minimum Risk Training}
Minimum Risk Training (MRT) \citep{Shen} aims to minimize the empirical risk of task loss over a larger set of $n=|\mathcal{S}|, n>1$ output samples $\mathcal{S}(x) \subset {Y}(x)$:
\begin{align*}
    \arg\min_{\theta} \sum_{(x,y')\in \mathcal{D}} \sum_{y\in \mathcal{S}(x)} {Q}_{\theta, \alpha}(y \mid x)[-\Delta(y, y')], \nonumber\\
    {Q}_{\theta, \alpha}(y \mid x^{(s)}) = \frac{p_{\theta}(y\mid x)^{\alpha}}{\sum_{y''\in \mathcal{S}(x)}p_{\theta}(y''\mid x)^{\alpha}}.
\end{align*}
As pointed out by~\citet{choshen}, MRT learns with biased stochastic estimates of the RL objective  
due to the renormalization of model scores, but that has not hindered its empirical success~\citep{Shen, edunov-etal-2018-classical, wieting-etal-2019-beyond, Wang2020OnEB}.
Interestingly, the resulting gradient update includes a renormalization of sampled rewards, yielding a similar effect to the baseline reward~\citep{Shen}. It also allows for more exploration thanks to learning from multiple samples per input, but it is therefore less attractive for human-in-the-loop learning and efficient training.

\subsection{Exposure Bias}
The \textit{exposure bias} in NMT arises from the model only being exposed to the ground truth during training, and receiving its own previous predictions during inference---while it might be overly reliant on perfect context, which in turn lets errors accumulate rapidly over long sequences~\citep{RanzatoETAL:16}. 
\citet{Wang2020OnEB} hypothesize that exposure bias increases the prevalence of hallucinations in domain adaptation and causes the \emph{beam search curse}~\citep{koehn-knowles-2017-six,yang-etal-2018-breaking}, which describes the problem that the model's performance worsens with large beams. \citet{Wang2020OnEB} find that MRT with multiple samples can mitigate this problem thanks to being exposed to model predictions during training. We will extend this finding to other PG variants with single samples.


\section{Experiments}
We implement PG and MRT (without enforcing gold tokens in $\mathcal{S}$; $n=5$) in Joey NMT~\citep{joeynmt} for Transformers \citep{vaswani}. We simulate rewards for training samples from IWSLT14 de-en with sacreBLEU~\citep{post-2018-call},
and test on IWSLT14 held-out sets. We consider two different domains for pretraining, WMT15 and IWSLT14.
 This allows us to distinguish the effects of RL in \emph{in-domain} learning vs \emph{domain adaptation} scenarios. 
RL experiments are repeated three times and we report mean and standard deviation. Remaining experimental details can be found in the Appendix. 
The goal is not to find the best model in a supervised domain adaptation setup (``Fine-tuning'' in Table~\ref{tab:domain_adapt}), but to investigate if/how scalar rewards expressing translation preferences can guide learning, mimicking a human-in-the-loop learning scenario.

\subsection{Peakiness}
\citet{choshen} suspect that PG improvements are due to an increase in peakiness. Increased peakiness is indicated by a disproportionate rise of $p_{top10}$ and $p_{mode}$, the average token probability of the 10 most likely tokens, and the mode, respectively. To test the influence of peakiness on performance, we deliberately increase and decrease the peakiness of the output distribution by adjusting the parameter $\tau$. In Tables~\ref{tab:in_domain} and ~\ref{tab:domain_adapt} we can see that all PG variants generally increase peakiness ($p_{top10}$ and $p_{mode}$), but that those with higher temperature $\tau>1$ show a lower increase. 
Comparing the peakiness with the BLEU scores, we find that BLEU gains are not tied to increasing peakiness in in-domain and cross-domain adaptation experiments.
This is exemplified by reward scaling (``PG+scaled''), which improves the BLEU but does not lead to an increase in peakiness compared to PG. These results show that improvements in BLEU can not just be explained by the peakiness effect. 
However, in cross-domain adaptation exploration plays a major role: Since the model is less familiar with the new data, reducing exploration (lower $\tau$) helps to improve translation quality.

\subsection{Upwards Mobility}
\begin{figure}
  \centering

  \includegraphics[width=1.\columnwidth]{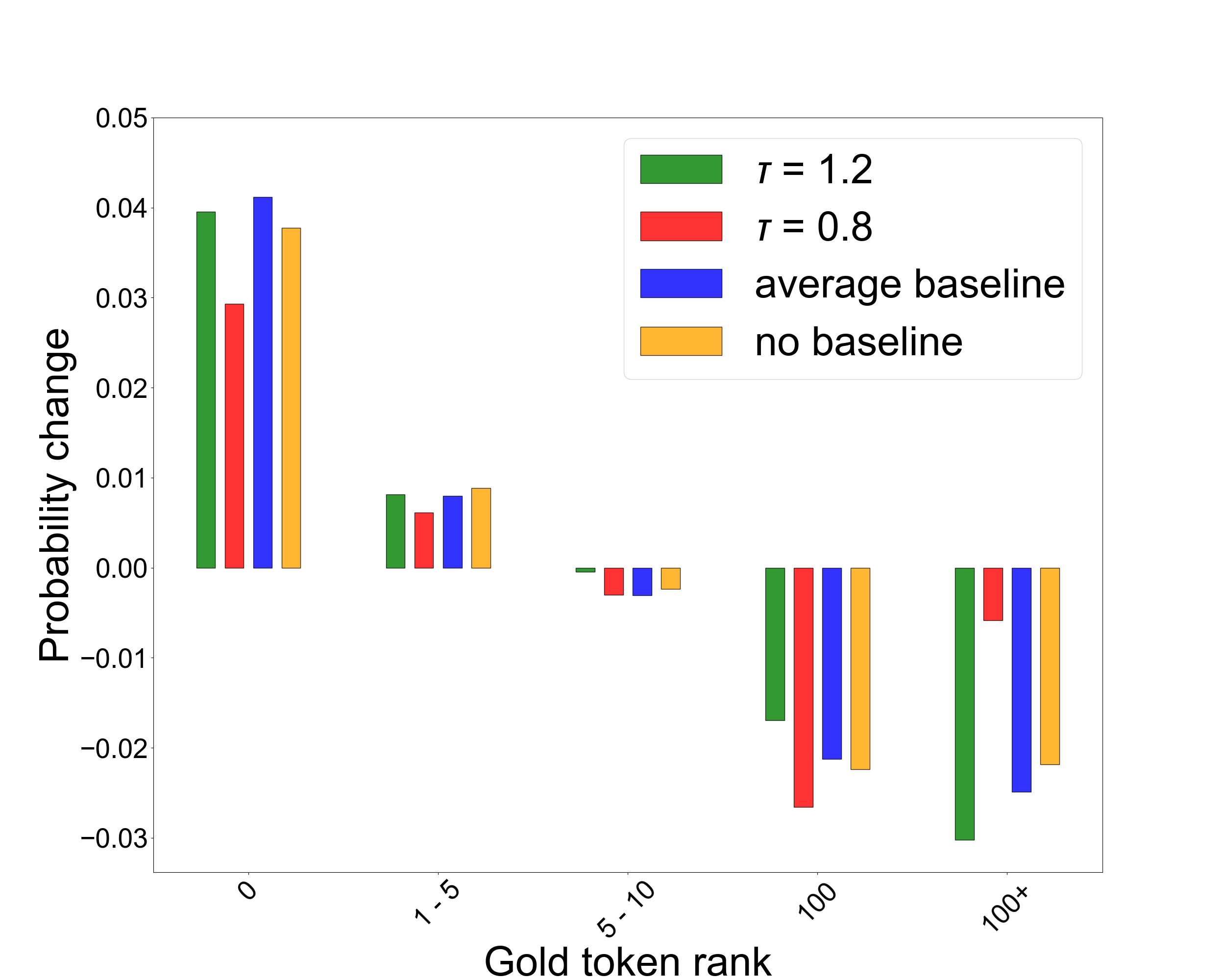}
  \caption{Change in probability for gold tokens to belong to each rank before and after RL on in-domain data.}
  \label{fig:no_bl_prob_change_domain_adapt_t=1}
\label{fig:probability_change_temperature}
\end{figure}

One disadvantage of high peakiness is that previously likely tokens accumulate even more probability mass during RL.~\citet{choshen} therefore fear that it might be close to impossible to transport lower-ranking tokens to higher ranks with RL. We test this hypothesis under different exploration settings by counting the number of gold tokens in each rank of the output distribution. That number is divided by the number of all gold tokens to obtain the probability of gold tokens appearing in each rank. We then compare the probability before and after RL. 
Fig.~\ref{fig:probability_change_temperature} illustrates that training with an increased temperature pushes more gold tokens out of the lowest rank. The baseline reward has a beneficial effect to that aim, since it allows down-weighing samples as well. This shows that upwards mobility is feasible and not a principled problem for PG.

\begin{table*}[t]
    \centering
    \begin{tabular}{l|cc|ccc}
        \toprule
 \textbf{Model}   & $\Delta p_{top10}$ & $\Delta p_{mode}$ & BLEU ($k=1$) & BLEU ($k=5$) & BLEU ($k=50$)\\
 \midrule
 Pretraining (WMT15) & 0 & 0 & 19.74 & 20.35  & 20.10 \\
 \midrule
 Self-training (IWSLT14)  & 46.87 & 99.15 &  $19.74\pm 0.00$ & $20.35\pm 0.00$ & $20.10 \pm 0.00$ \\ 
  Fine-Tuning (IWSLT14) & 16.71 & 25.29 & $28.25 \pm0.11 $ & $29.38 \pm 0.09$ & $29.48 \pm 0.05$\\
 \midrule
PG ($n=1$) & 44.10 & 88.60 & $22.34  \pm 0.73$ & $22.62 \pm 0.56$ & $22.60 \pm 0.51$ \\ 
 PG + scaled  & 43.46  & 87.83 & $23.23 \pm 0.12$  & $23.50 \pm 0.15$ & $23.52 \pm 0.19$ \\
 PG + average bl  & 44.91 &  94.65 & $24.31 \pm 0.41$  & $24.53 \pm 0.26$ & $24.56 \pm 0.22$ \\ 
 PG + $\tau = 1.2$   & 39.05  & 78.93 & $21.26 \pm 0.20$ & $21.58 \pm 0.21$ & $21.60 \pm 0.28$\\ 
 PG + $\tau = 0.8$  & 46.91  & 95.10 & $23.27 \pm 0.24$ & $23.82 \pm 0.22$ & $23.93\pm 0.23$ \\
 PG + constant  &  48.50 & 115.31 & $19.74 \pm 0.00$ & $20.35 \pm 0.00$ & $20.10 \pm 0.00$\\
 \midrule
 PG + average bl + $\tau = 1.05$   & 44.44  & 93.53 & $24.18 \pm 0.09$ & $24.53 \pm 0.09$ & $24.60 \pm 0.11$\\ 
 PG + average bl + $\tau = 0.95$  & 45.34  & 95.04  & $24.41 \pm 0.30$ & $24.86 \pm 0.17$ & $24.86\pm 0.11$ \\
 \midrule
MRT ($n=5$) & 44.63 & 103.68 & $26.98 \pm 0.10$  & $27.08 \pm 0.09$ & $27.09 \pm 0.10$ \\
\bottomrule
    \end{tabular}
    \caption{Cross-domain adaptation: Peakiness indicators (\%), and IWSLT14 test set results for beam size $k$.}
    \label{tab:domain_adapt}
\end{table*}

\subsection{Meaningful Rewards}
\citet{choshen} observe an increase in peakiness when all rewards are set to 1, and BLEU improvements even comparable to BLEU rewards. 
While our results with a constant reward of 1 (``PG+constant'') also show an increase in peakiness for cross-domain adaptation (Table \ref{tab:domain_adapt}), we observe deteriorating scores using the same hyperparameters as with other PG variants. Reducing the learning rate from $1\times10^{-4}$ to $1\times10^{-5}$ alleviates these effects but still does not lead to notable improvements over the pretrained model (with an average gain of 0.09 BLEU). 
Similarly, domain adaptation via self-training increases peakiness but does not show improvements over the baseline. A smaller learning rate of $1\times10^{-5}$ reduces peakiness and shows a similar behaviour as constant rewards (a gain of 0.1 BLEU).
These results confirm that gains do not come from being exposed to new inputs and increased peakiness, which contradicts the results of~\citet{choshen}. 
While the effects in-domain are generally weak with a maximum gain of 0.5 BLEU over the baseline (with beam size $k=5$, Table~\ref{tab:in_domain}), the results for domain adaptation (Table~\ref{tab:domain_adapt}) show a clear advantage of using informative rewards with up to +4.5 BLEU for PG and +6.7 BLEU for MRT (with beam size $k=5$). We conclude that rewards do matter for PG for NMT. 

\subsection{Allowing Negative Rewards}
As described in Section \ref{sec:modidied rewards}, scaling the reward 
(``PG+scaled''), subtracting a baseline (``PG+average bl''), or normalizing it over multiple samples for MRT, introduces negative rewards, which enables updates away from sampled outputs.
BLEU under domain shift (Table \ref{tab:domain_adapt}) shows a significant improvement when allowing negative rewards. 
The scaled reward increases the score by almost 1 BLEU, the average reward baseline by almost 2 BLEU and MRT leads to a gain of about 4.5 BLEU over plain PG. 

\subsection{The Beam Curse}

The results show that improvements of RL over the baseline are higher with lower beam sizes, since RL reduces the need for exploration (through search) during inference thanks to the exploration during training.
These findings are in line with \citet{ac_bahdanau}. For RL models, BLEU reductions caused by larger beams are weaker than for the baseline model in both settings, which confirms that PG methods are effective at mitigating the beam search problem, and according to \citet{Wang2020OnEB} might also reduce hallucinations. 

\subsection{Discussion}
Despite the promising empirical gains over a pretrained baseline, all above methods would fail if trained from scratch, as there are no non-zero-reward translation outputs sampled when starting from a random policy. Empirical improvements over a strong pretrained model vanish when there is little to learn from the new feedback, e.g. when it is given on the same data which the model was already trained on, as we have shown above, relating to the ``failure'' cases in \citet{choshen}. RL methods for MT can be effective at adapting a model to new custom preferences if these preferences can be reflected in an appropriate reward function, which we simulated with in-domain data. In Table~\ref{tab:domain_adapt}, we observed this effect and gained several BLEU points without revealing reference translations to the model. Being exposed to new sources alone (without rewards) is not sufficient to obtain improvements, which we tested by self-training (Table~\ref{tab:domain_adapt}). 
Ultimately, the potential to improve MT models with RL methods lies in situations where there are no reference translations but reward signals, and models can be pretrained on existing data.

\section{Conclusion}
We provided empirical counter-evidence for some of the claimed weaknesses of RL in NMT by untying BLEU gains from peakiness, showcasing the upwards mobility of low-ranking tokens, and re-confirming the importance of reward functions. The affirmed gains of PG variants in adaptation scenarios and their responsiveness to reward functions, combined with exposure bias repair and avoidance of the beam curse, rekindle the potential to utilize them for adapting models to human preferences.

\section*{Acknowledgements}
We acknowledge the support by the state of Baden-Württemberg through bwHPC compute resources.

\bibliography{anthology,custom}
\bibliographystyle{acl_natbib}

\clearpage
\appendix
\section{Data}\label{app:data}
\begin{table}[h]
    \centering
    \begin{tabular}{lrrr}
    \toprule
       \textbf{ Domain} & \textbf{Train} & \textbf{Dev} & \textbf{Test} \\
        \midrule
        WMT15 & \num{3898886} & \num{8496} & \num{2902} \\
        IWSLT14 & \num{159392} & \num{7245} & \num{6750} \\
        \bottomrule
    \end{tabular}
    \caption{Training, dev, and test sizes for WMT15 and IWSLT14 de-en data in number of sentences.}
    \label{tab:data}
\end{table}
Table~\ref{tab:data} lists the sizes of data split for the parallel datasets from WMT15~\citep{bojar-EtAl:2015:WMT} and IWSLT14\footnote{\url{https://sites.google.com/site/iwsltevaluation2014/mt-track}} used in the experiments. The two datasets are preprocessed using scripts from the Moses toolkit.\footnote{\url{https://github.com/moses-smt/mosesdecoder/tree/master/scripts}} The preprocessing pipeline contains the following steps: 
\begin{itemize}
    \item Tokenization with \texttt{tokenizer.perl} 
    \item Lowercasing with \texttt{lowercase.perl}
    \item Filtering using \texttt{clean-corpus-n.perl}. Sentences with more than 80 words are removed from the dataset
\end{itemize}
Additionally, we applied Byte-Pair-Encoding~\citep{sennrich-haddow-birch:2016:P16-12} using subword-nmt\footnote{\url{https://github.com/rsennrich/subword-nmt}} to create subword units. 

\section{Model Configurations}\label{app:configurations}
Table~\ref{tab:pretrain_hyperparams} contains the hyperparameters as Joey NMT configurations for the pretrained models, Table~\ref{tab:PG_hyperparams} the modified hyperparameters for PG, and Table~\ref{tab:MRT_hyperparams} the modified hyperparameters for MRT. 
Random seeds for the three runs were set to 42, 8 and 64.

\section{Sample Efficiency}\label{app:cost-effectiveness}
\begin{figure}
  \centering
  \includegraphics[width=1\columnwidth]{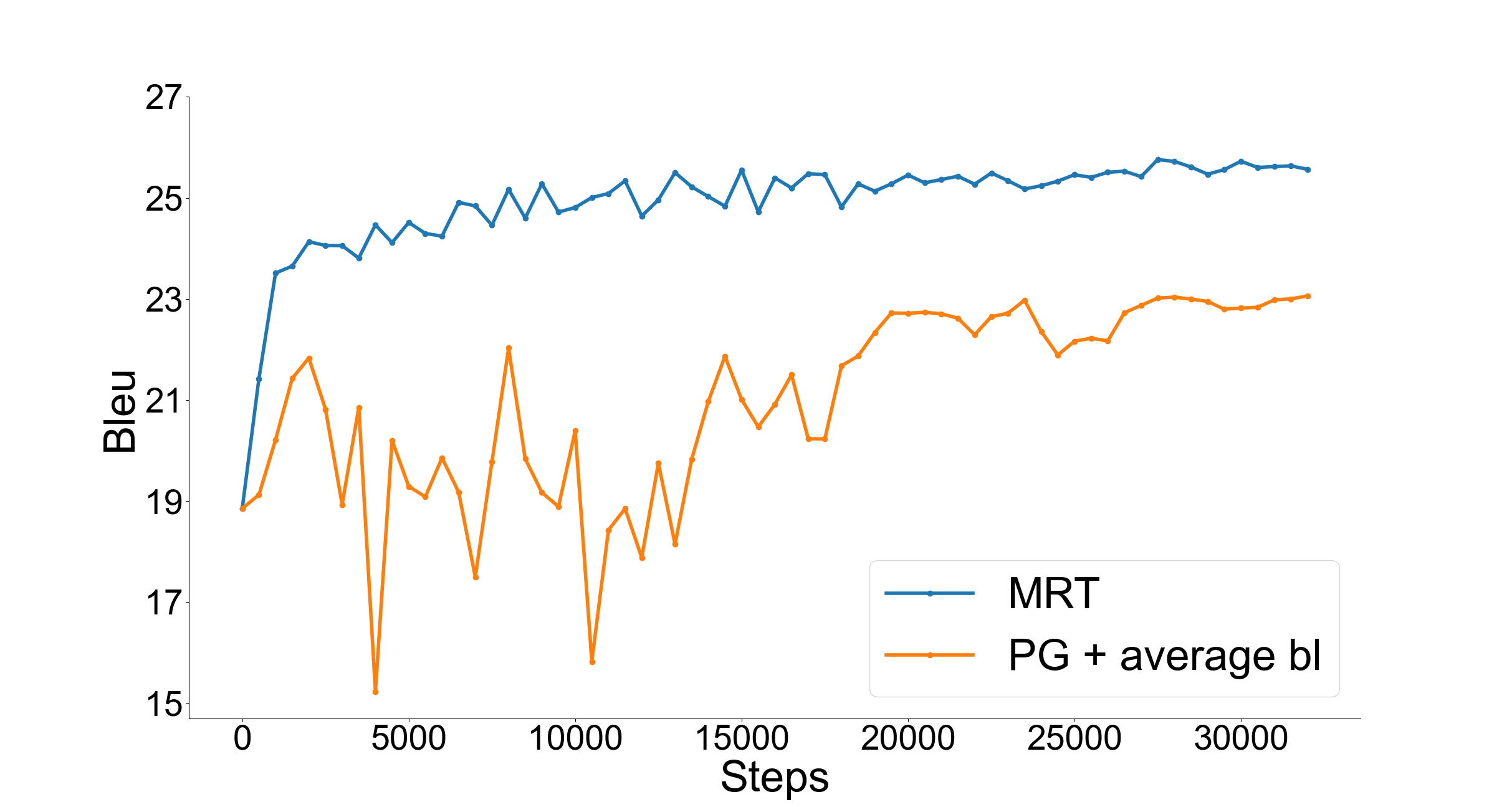}
  \caption{Dev BLEU/steps for cross-domain.}
\label{fig:da_steps}
\end{figure}

Fig.~\ref{fig:da_steps} shows that both MRT and Policy Gradient need a comparable amount of steps (with a batch size of 256 tokens) to reach their optimum. However, the performance of MRT is more stable over the course of training, while Policy Gradient shows higher variance. MRT learns from $n=5$ outputs and rewards per step (compared to Policy Gradient with $n=1$), which stabilizes the updates.   

\section{Additional Considerations}\label{app:considerations}
\paragraph{Learned Baseline} A reward baseline can also be learned by formulating it as a regression problem, but like \citet{wu-etal-2018-study} we found no empirical gains, thus excluded it from the experiments reported in this paper.

\paragraph{Scaling Rewards}
We found that selecting $max$ and $min$ over all previous rewards led to deteriorating BLEU scores. This is why we recompute them for each batch.

\paragraph{Gold Tokens in MRT} 
\citet{Shen} add the gold sequence to the sample space. However, \citet{edunov-etal-2018-classical} find that this destabilizes training, so \citet{choshen} and \citet{Wang2020OnEB} choose to omit it, and so do we.

\section{Development Results}\label{app:dev}
Tables~\ref{tab:all algorithms in domain} and~\ref{tab:all algorithms domain adaptation} report results on the development set that were used for tuning the models. 
They show stable results across different held-out sets.

\section{Absolute Peakiness}\label{app:peakiness}
Tables~\ref{tab:peakiness_in_domain_absolute_values} and ~\ref{tab:peakiness_domain_adapt_absolute_values} contain the absolute values for the change in peakiness that were used to compute percentages for the main paper results.

\begin{table}[]
    \centering
    \begin{tabular}{ m{2.5cm} m{2cm} m{2cm} }
    \toprule
    \textbf{Model} & \textbf{Dev BLEU} & \textbf{Test BLEU}  \\ 
    \midrule
    Pretraining &  34.26 & 34.12  \\ 
    \midrule
    PG & $34.37 \pm 0.12$ & $34.24 \pm 0.04$  \\ 
    PG + average bl & $34.75 \pm 0.06$ & $34.40 \pm 0.03$ \\ 
    PG + $\tau = 1.2$ & $34.46 \pm 0.10$ & $34.15 \pm 0.02$ \\ 
    PG + $\tau = 0.8$ & $34.47 \pm 0.12$ & $34.26 \pm 0.11$  \\
    PG + constant & $34.26 \pm 0$ & $34.13 \pm 0.01$  \\
    PG + scaled & $34.55 \pm 0.10$ & $34.30 \pm 0.10$  \\
    \bottomrule
    \end{tabular}
    \caption{PG variants in-domain adaptation (IWSLT14), beam size=5.}
    \label{tab:all algorithms in domain}
\end{table}

\begin{table}[]
    \centering
    \begin{tabular}{ m{2.5cm} m{2cm} m{2cm} }
    \toprule
    \textbf{Model} &\textbf{ Dev BLEU} & \textbf{Test BLEU}  \\ 
    \midrule
    Pretraining &  18.86 & 20.35  \\ 
    \midrule
    PG & $21.63 \pm 0.39$ & $22.62 \pm 0.56$  \\ 
    PG + average bl & $23.31 \pm 0.32$ & $24.53 \pm 0.26$ \\ 
    PG + $\tau = 1.2$ & $20.45 \pm 0.18$ & $21.58 \pm 0.21$ \\ 
    PG + $\tau = 0.8$ & $22.42 \pm 0.08$ & $23.82 \pm 0.22$  \\ 
    PG + constant & $18.86 \pm 0$ & $20.35 \pm 0$  \\
    PG + scaled & $22.40 \pm 0.10$ & $23.50 \pm 0.15$  \\
    \bottomrule
    \end{tabular}
    \caption{PG variants cross-domain adaptation (WMT15 to IWSLT14), beam size=5}
    \label{tab:all algorithms domain adaptation}
\end{table}

\begin{table}[]
\begin{center}
\begin{tabular}{ m{2.5cm} m{1.2cm} m{1.2cm} m{1.2cm} } 
\toprule
\textbf{ Model }  & $\Delta p_{top10}$ & $\Delta p_{mode}$ & $\Delta p_{gold}$  \\ 
\midrule
PG  & $0.294 \pm 0.001$ & $0.400 \pm 0.001$ & $0.095 \pm 0.002$ \\ 
PG + scaled  & $0.290 \pm 0.002$  & $0.397 \pm 0.006$ & $0.098 \pm 0.001$ \\
PG + average bl  & $0.300 \pm 0.001$ & $0.428 \pm 0.001$ &  $0.103 \pm 0.002$ \\ 
PG + $\tau = 1.2$   & $0.260 \pm 0.003$  & $0.357 \pm 0.006$ &  $0.082 \pm 0.001$ \\
PG + $\tau = 0.8$  & $0.313 \pm 0.001$  & $0.430 \pm 0.001$ &  $0.102 \pm 0.001$ \\
PG + constant  &  $0.323 \pm  0.001$ & $0.521 \pm 0.002$ & $-0.123 \pm 0.004$ \\
\midrule
MRT & $0.307 \pm 0.002$ & $0.482 \pm 0.004$ & $0.126 \pm 0.004$ \\
\bottomrule
\end{tabular}
\caption{Absolute changes in peakiness for cross-domain adaptation (WMT15 to IWSLT14).}

\label{tab:peakiness_domain_adapt_absolute_values}
\end{center}
\end{table}

\begin{table}[]
\begin{center}
\begin{tabular}{ m{2.5cm} m{1.2cm} m{1.2cm} m{1.2cm}} 
\toprule
 \textbf{Model}  & $\Delta p_{top10}$ & $\Delta p_{mode}$ & $\Delta p_{gold}$  \\ 
\midrule
PG  & $0.099 \pm 0.002$ & $0.199 \pm  0.003$ & $0.076 \pm 0.001$ \\ 
PG + scaled  & $0.097 \pm 0.002$  & $0.162 \pm 0.005$ & $0.074 \pm 0.002$  \\
PG + average bl  & $0.108 \pm  0.002$ &  $0.185 \pm 0.006$ & $0.083 \pm 0.002$ \\
PG + $\tau = 1.2$  & $0.052 \pm 0.002$ & $0.112 \pm  0.004$ & $0.060 \pm 0.001$ \\
PG + $\tau = 0.8$  & $0.123 \pm 0.001$ & $0.198 \pm 0.003$ & $0.083 \pm 0.003$ \\
PG + constant  & $0.012 \pm 0.007$ & $0.017 \pm 0.007$ & $0.003 \pm  0.001$  \\
\midrule
MRT & $0.112 \pm 0.003$ & $0.217 \pm 0.004$ & $0.090 \pm 0.004$ \\
\bottomrule
\end{tabular}
\caption{Absolute changes in peakiness for in-domain (IWSLT14) adaptation.}
\label{tab:peakiness_in_domain_absolute_values}
\end{center}
\end{table}

\begin{table}[]
\begin{center}
\resizebox{\columnwidth}{!}{%
\begin{tabular}{  m{4cm}  m{2.2cm} m{2.2cm} }
\toprule
\textbf{Model} & \textbf{IWSLT14} & \textbf{WMT15} \\
\midrule
Parameter & Setting  & Setting \\ 
\midrule
initializer & "xavier" & "xavier"\\
embed initializer &  "xavier" & "xavier"\\
embed init gain &  1.0 &  1.0\\
init gain &  1.0 &  1.0\\
bias initializer & "zeros"& "zeros"\\
tied embeddings &  True&  True\\
tied softmax &  True&  True\\

encoder type & Transformer & Transfomer\\ 
encoder embeddings dim & 256 & 128  \\ 
encoder hidden size & 256  & 128\\ 
encoder dropout & 0.3 &  0.3\\
encoder num layers & 6 & 6\\
encoder num heads & 4  & 4\\
encoder ff\_size & 1024  & 512 \\ 
\midrule
decoder type & Transformer & Transformer \\ 
decoder embeddings dim & 256 & 128 \\ 
decoder hidden size & 256  & 128\\ 
decoder dropout & 0.3  & 0.3 \\
decoder num layers & 6 &  6\\
decoder num heads & 4 & 4\\
decoder ff\_size & 1024 & 512 \\
\midrule
optimizer & "adam" & "adam" \\
normalization &  "tokens" &  "tokens"\\
adam\_betas& [0.9, 0.999] & [0.9, 0.999] \\
scheduling& "plateau" & "plateau" \\
patience& 5 & 5 \\
decrease\_factor& 0.7 & 0.7 \\
loss& "crossentropy" & "crossentropy" \\
learning rate& 0.0003 & 0.0003 \\
learning rate\_min& 0.00000002 & 0.00000002 \\
weight decay& 0.0 & 0.0 \\
label smoothing& 0.1 & 0.1 \\
batch size& 2048 & 4096 \\
batch type& "token" & "token" \\
epochs& 100 & 100 \\
\bottomrule
\end{tabular}%
}
\caption{Pretraining model parameters}
\label{tab:pretrain_hyperparams}
\end{center}
\end{table}

\begin{table}[]
\begin{center}
\begin{tabular}{  m{2.6cm}  m{1.8cm} m{2.2cm} }
\toprule
Parameter & In-domain & Cross-domain   \\ 
\midrule
learning rate & 0.00001 & 0.0001\\
batch size & 128 & 256 \\
\bottomrule
\end{tabular}%
\caption{Policy Gradient parameters}
\label{tab:PG_hyperparams}
\end{center}
\end{table}

\begin{table}[]
\begin{center}
\begin{tabular}{  m{2.6cm}  m{1.8cm} m{2.2cm} }
\toprule
Parameter & In-domain & Cross-domain  \\ 
\midrule
learning rate & 0.00001 & 0.0001\\
batch size& 32 & 64\\
batch multiplier& 4 & 4\\
eval batch size & 128 & 128\\
samples & 5 & 5 \\
alpha & 0.005 & 0.005\\
\bottomrule
\end{tabular}%
\caption{MRT parameters}
\label{tab:MRT_hyperparams}
\end{center}
\end{table}

\end{document}